\documentclass{article}
\title{The Movie Graph Argument Revisited}
\author{Russell Standish\\Economics\\Kingston University}

\newtheorem{definition}{Definition}
\newtheorem{remark}{Remark}

\begin{document}
\maketitle

\begin{abstract} In this paper, we reexamine the {\em Movie Graph
Argument}, which demonstrates a basic incompatibility between
computationalism and materialism. We discover that the incompatibility
is only manifest in singular classical-like universes. If we accept
that we live in a Multiverse, then the incompatibility goes away, but
in that case another line of argument shows that with
computationalism, the fundamental, or primitive materiality has no causal
influence on what is observed, which must must be derivable from basic
arithmetic properties.
\end{abstract}

\section{Introduction}

Computationalism is the idea that our minds are computational
processes, and nothing but. In particular, an appropriate program
running on a computer will instantiate consciousness just as well as
brains made of neurons. Bruno Marchal, who developed the Movie Graph
Argument\cite{Marchal88} is fond of introducing this concept via a
parable: 
\begin{quote}
You have just discovered you have terminal brain cancer, and
the doctor treating you proposes replacing your brain by an electronic computer
running an artificial intelligence program initialised by the synaptic
weights read out from your old brain prior to its destruction.
\end{quote}

Would you say ``yes'' to the doctor? Do you think you will survive the
transplant? If not, then what if the doctor proposes replacing your
brain with a detailed emulation, including chemical and electrical
properties, of all of the atoms making up your brain? If not, then
what if the proposal were a detailed quantum mechanical simulation of
the elementary particles making up your atoms.

If you say yes at any point, you are affirming computationalism.

However, computationalism implies a number of surprising
consequences. Because it is easy to copy a computer program, it should
be possible to be cloned into a doppelg\"anger, whose memories are
identical to one's own up to the point of being cloned. The computer
program can be transferred over the internet, allowing teleportation.

Moreover, it is possible to reimplement the exact same program in
different ways, for example by coding the program using different
programming languages. The detailed use of a machine's registers, and
instructions executed will differ dramatically in each case, yet the
conscious experience should be identical if the program is implemented
correctly. Furthermore, if the original program is run again with the
exact same input, the same conscious experience will be generated, and
the exact same sequence of register states will be activated. A
recording of this sequence of register states when played back will be
physically indistinguishable from the running of the original
computation. The Movie Graph Argument seeks to parlay this into an
absurdity, whereby if the physical activity were all that was
important for consciousness, we would need to attribute consciousness
or not to a physical system, only if the system were observed. One is
left to conclude that computational and physical supervenience are
fundamentally incompatible concepts.

Before introducing the Movie Graph Argument, I'd like to introduce a
number of prerequisite topics, the {\em Universal Dovetailer
Argument}, the concept of supervenience, and Tim Maudlin's Olympia argument.

\section{Universal Dovetailer Argument}

The {\em universal dovetailer} is a computer program invented by
Marchal\cite{Marchal91} that effectively executes all possible
computer programs, on all possible inputs, albeit with exponential
slowdown. It works by executing the first step of the first program,
then the first step of the second program, the second step of the
first, second of the second, first of the third, and so on,
zig-zagging between executing the next step and starting a new program.

Clearly, if computationalism is valid, then all possible experiences
are instantiated by the dovetailer. Each experience will occur with a
certain measure within the dovetailer, a measure moreover that is
independent of the specific universal machine, or the specific
universal dovetailer used (see appendix). What we experience will be
drawn randomly from those experiences, hence typically one with high
measure in the dovetailer.

One of the consequences of the universal dovetailer argument is that
you cannot tell which computer program is you. For every program that
instantiates your current conscious state, there are an infinite
number of other possible programs that generate the same history of
conscious states to that point in time, but differ after that time,
corresponding to different possible futures. This leads to an
irreduciable indeterminism --- even an omniscient god cannot know what
you will experience next. The question is ill-posed --- which of the
possible future ``yous'' is the real one?

This indeterminism, which Marchal calls {\em first person
indeterminancy} (FPI)\cite{Marchal04} is related strongly to how quantum
indeterminism appears within the deterministic Many Worlds
Interpretation of quantum mechanics.

This indeterminism also implies we are made up of all computations
having the same initial history that computes our current conscious
state. This will feature in the definition of the {\em computational
supervenience thesis} in \S\ref{MGA}.

However, the Church-Turing thesis implies that it doesn't matter what
physical computer the dovetailer is run on. It could equally be a
contraption of gears and cogs, like Babbage's analytic engine, or
pebbles on a lattice with an infinitely patient child moving them
according to the rules of the {\em Game of Life}, as an electronic
computer we know today. Our experienced physical world must therefore
be independent of any such such primitive physical substrate. The
supposition that the computations instantiating our minds run on a
particular piece of hardware serves no explanatory purpose
whatsoever. All physical experiences are grounded in the properties of
the Universal Dovetailer running on a Universal Machine, which is a
purely mathematical notion. We might describe this consequence of the
Church-Turing thesis as a Turing curtain, that prevents us from ever
accessing ontological realty.

The {\em Universal Dovetailer Argument}\cite{Marchal04} steps 1--7
presents this radical conclusion that the phenomenally observed
physical world cannot be ontologically primitive if computationalism
is true, in a series of steps that build gradually upon the reader's
intuition.

The wrinkle is to suppose that the ontological universe doesn't have
sufficient resources to run a universal dovetailer, in which case it
really does matter how powerful the hardware is running our
reality. Whilst a universal dovetailer can be coded and run on the
physical computers we have today, in practice only a short initial
portion of the UD can be run. What if the universe goes into a heat
death before any conscious program is started? In such a case,
properties of the ontological world leak through the the Turing
curtain --- ontological physics is reflected in phenomenal physics. It
should be pointed out that if this were true, then there must be a
largest number that is computed during the universe's history, a
philosophy of mathematics going by the name of {\em
ultrafinitism}\cite{zeilberger01} or {\em strict finitism}\cite{Tiles89}.

To distinguish between these cases, Marchal calls a universe capable
of running a universal dovetailer fully a {\em robust}
universe. Whilst such a universe is necessarily infinite, we can, for
the purposes of this argument, consider robust universes to be ones
that can run enough of the universal dovetailer for programs
instantiating all possible human experiences of consciousnesses within a
human life time be executed. This is still an immense universe, but not
necessarily an infinite one. 

Since all our possibe experiences will be instantiated, and observed,
our phenomenal physics depends only on the properties of the universal
machine, not on any underlying physical sustrate.  The only phenomenal
impact will be a subtle departure from the universal dovetailer
measure.

A non-robust universe is incapable of generating consciousness by
running the universal dovetailer. Conscious entities can only appear
by a correct program being instantiated at a primitive physical level,
whilst other experiences and entities are not so instantiated. The
primitive physical world then potentially has a causative effect on
phenomenal physics, for example by allowing some experiences to be
experienced, and not others.

The Movie Graph Argument (MGA) was developed to show that even in non-robust
universes, primitive physics plays no explanatory role. But before
presenting the MGA, we must first introduce the notion of {\em
supervenience}, and also discuss a rather similar argument by Tim
Maudlin, pointing at the inconsistency of computationalism with materialism.

\section{Supervenience}

{\em Supervenience} is an attempt to capture the dependence of some
phenomenon on its substrate\cite{McLaughlin-Bennett14}. Loosely
speaking, a phenomenon {\em supervenes} on a substrate if a change in
the supervening phenomenon ncessarily entails a change in the
substrate. For example, consider the phenomenon of speech. The
different situations where the spoken words ``hello'' and ``hi'' are
uttered, will necessarily involve different motions of the air
molecules, so we can say that speech supervenes on molecular motion.

In the case of consciousness, it is widely believed that consciousness
supervenes on our brains, as it is observed that different conscious
experiences are accompanied by different brain states.

Now consider the scenario of a class of school children, one of whom
is named Alice, and another Bob. Does Alice's consciousness supervene
on the class? Well, yes, as we observe that any change in Alice's
consciousness must correspond to a physical change in the classroom,
concentrated in Alice's brain. But we can ask a slight different
question --- does consciousness supervene on the class. In this case,
we'd have to answer no, because both Alice's conscious states and
Bob's, not to mention the teacher's and other students are all present
in the class. A difference in conscious state does not correspond to a
physical difference, but merely to an indexical change: which
consciousness we're referring to. We can express the same conundrum
using the speech case, exploiting the so-called ``cocktail party''
effect. Alice says ``hello'', and Bob says ``hi'' simultaneously ---
but which word we hear depends on who we're actively listening to at
the time. The words no longer supervene on the air molecules, but on
the state of the listener.

To see how this applies to the universal dovetailer, recall that the
universal dovetailer instantiates all possible experiences. A
different experience does not entail a difference in the universal
dovetailer. So counterintuitively, consciousness cannot supervene on
the universal dovetailer itself, even though according to
computationalism, it will supervene on some of the non-dovetailing
computations being executed by the dovetailer.

\section{Computational Supervenience and Counterfactual Equivalence}

The basic idea of the computational supervenience thesis is that a
conscious state supervenes on a computation. Of course there are many
running programs that perform the same computation. The most trivial
example of these being programs that perform the same steps up to some
time $t$, but then diverge after that time, which is the source of the
first person indeterminism. But it is also true that two distinct
programs can pass through the same sequence of machine states,
without being computationally equivalent. 

The simplest example of such a difference might be if program A
executes the ``or" instruction on registers $x$ and $y$, and B
executes the ``and" instruction.  If it so happens that both $x$ and
$y$ both contain the same value (both true or both false), then the
resultant machine state is identical with each program. Yet the two
programs are quite different, as if the two registers had different
values, the resulting machine state would be quite different. We call
this ``if it had been different'' a {\em counterfactual}. In this
case, programs A and B are not counterfactually equivalent.

Computational supervenience entails that two counterfactually
equivalent programs instantiate the same conscious experience, but is
mute on whether two counterfactually inequivalent programs that happen
to pass through the same sequence of machine states instantiate the
same conscious experiences. On the other hand, physical supervenience,
required by materialism, asserts that two machines passing through the
same states must instantiate the same conscious experience.

The heart of the Movie Graph Argument, and also of Tim Maudlin's is
to set up an absurdity, where a very simple computation that seems
unlikely to be instantiating consciousness, nevertheless only differs
physically from a conscious computation by mostly physically inactive
components. 

%It seems plausible that counterfactual inequivalence is needed as part
%of the definition of what could differ between computations supporting
%different conscious experiences. This is supported by the intuition
%that a mere playback of a recording of a conscious machine (eg
%reanimating a dead brain by passing recorded EEG sgnals through the
%neurons) is not sufficient to instantiate a consciousness. A recording
%playback (when done perfectly) will pass through the same sequence of
%machine states as the original conscious computation, yet it need not
%be conscious because it is not {\em counterfactually equivalent} to
%the original computation (we would say it is not {\em counterfactually
%correct}). 

\section{Maudlin's Olympia Argument}\label{Olympia}

Tim Maudlin presented an interesting argument that computationalism is
incompatible with materialism\cite{Maudlin89}, which he defines as a form
of physical supervenience --- that consciousness supervenes on
physical activity. To summarise his argument, he transforms the
physical process performing a conscious computation into one replaying
a recording of the process. In a nod to Hoffman's tale {\em Der
Sandmann}, Maudlin calls the former machine Klara and the latter
Olympia. The machinery passes through the exact same sequence of
states in both cases, but clearly in the second case the computation
is utterly trivial --- reading the machine state from a recording.
The unstated assumption is that Olympia is too simple to be conscious.

It might be objected that Olympia is not counterfactually
correct. Klara performs the calculation, and so would produce a
different result if some of the intermediate results
differed. Olympia, on the other hand, knows ahead of time what the
states of the registers are. If the registers were different, it would
still produce the same sequence of states as specified in the
recording.

To counter this objection, Maudlin introduces a baroque construction
of attaching a copy of Klara to each and every state of the
sequence. Each Klara has been advanced to the point in calculation
corresponding to the step to which it is attached. If the intermediate
result differs (not that it will) from that of the recording at some
step, the attached Klara will take over the computation from that
point, thus preserving counterfactual correctness. Yet, these Klaras
are physically inert, as these counterfactual states never
occur. Maudlin's point is that if counterfactual correctness is
relevant, then a simple switch connecting the physically inert Klaras to Olympia
suffices to switch consciousness on and off.

An interesting objection to Maudlin's construction was pointed out by
Colin Klein\cite{Klein15}, in which he notes that Olympia is a special case of a
more powerful machine called an oracle machine, where the oracle
consists of the output of a Turing computation, and so shouldn't
really be considered the sort of computation that consciousness might
supervene on.

\section{Multiverse objection to Maudlin's argument}

If the Many Worlds Interpretation of quantum mechanics is literally
true, then we must also consider that the counterfactuals will occur in
alternate universes. In Maudlin's setup, the program either has fixed
inputs, corresponding to a specific history, or has no inputs, perhaps
corresponding to a dreaming state. If the situation is one of fixed
inputs, then it is easy to see that the quantum multiverse must also
contain versions of the same program with differing inputs. If it is
the no input situation, then counterfactual states must also occur in
any physical implementation due to the possibility of error or noise
in the implementation.

So if counterfactual situations are physically realised somewhere in
the multiverse, then one can no longer claim that the attached Klaras
in Maudlin's thought experiment are physically inert.

Nevertheless, this objection is not a valid objection to the use of
the Maudlin's argument, nor the MGA for step 8 of the UDA to obtain the
incompatibility of computationalism and materialism for non-robust
universes. The reason is that a multiverse is a physical quantum
computer, and at least all possible human experiences are experienced somewhere
in the Multiverse, if not all possible conscious experiences. Thus the
Multiverse, even if finite in size, should be considered a robust universe.

A slightly different corollary that sidesteps the issue of whether the
Multiverse is real is to conclude that conscious states can only occur
if the conscious computation is accepting inputs from a physical
environment that is non-computational. The physical environment
observed by conscious being instantiated by the universal dovetailer
is non-computational, precisely because of the presence of randomness
induced by first person indeterminancy. Non input computations of the
sort Maudlin uses in his argument, sometimes described as ``dreaming
states'', cannot be conscious. 

It should be pointed out that when an animal dreams, the brain
is processing stochastic data generated within the brain, such as
synaptic noise amplified by chaotic dynamic system effects. So to
describe a dream as the running of a non-input program is an
unjustifiable assumption, and consequently the necessity of a
stochastic environment can be compatible with the conscious experience
of dreams.

\section{The Movie Graph Argument}\label{MGA}

Marchal originally presented his argument in French as {\em l'argument
du graphe film\'e}\cite{Marchal88}, literally {\em the filmed graph
argument}, which got rendered into English as the Movie Graph
Argument. The idea is that the conscious computation is implemented as
a graph (or network) of stateful objects (eg abstract neurons)
embedded in a glass plate, exchanging photons for messages. This
allows a movie camera to record a movie of the operation of the
artificial brain. Then by parts, he severs some of the network links
between neurons, but by projecting the movie back onto the network, is
able to excite those neurons as though they were still connected. The
result, like Maudlin's argument, is a physically identical process (at
the state level) that is however, computationally not identical. The
resulting system consists of two parts, the original glass plate,
which is now basically inert, and the projector of the movie. Since
the glass plate now has no causal effect, it can be exised from the
system, leaving the movie. But then the light illuminating the film
has no causal role, and can be switched off. Finally, one is left with
a film moving through the pellicule. But motion is relative, so this
is equivalent to the film being stationary, and the observer being in
motion. So no physical supervenience entails that the entitiy is
conscious if and only if an observer is in relative motion to the film
--- an absurdity indeed.

Where this argument appears to break down is where the film comes from
in the first place. In the thought experiment, the original physical
process supporting the conscious moment is filmed or otherwise
recorded. In replaying the recording, the conscious moment is not
changed in any way. The physical environment observed by the conscious
moment remains unchanged too. If the conscious moment differed, then
the recording, if any, would differ, however absence of a recording
does not diminish the conscious experience. Thus there is a partial
supervenience of the consciousness on the recording. To subvert this
line of thinking, Marchal talks of physical supervenience at a time
and place. Even though the replaying of the recording is physically
identical to the earlier computation (aside from some additional
machinery that merely extends the physical system), it's presence at a
time and place does not attach that conscious experience to that time
and place. The conscious experience supervenes only on the original
computational process.

\section{Conclusion}

Physical supervenience is simply not compatible with computational
supervenience in a non-robust universe. In order for physical
supervenience to be compatible with computational supervenience, we
need to inhabit a Multiverse, which as noted above is a robust case.

The anthropic principle is the notion that our physical environment is
compatible with our physical existence within that environment. This
is a form of physical supervenience, and observed evidence consistently
points to the anthropic principle as being true.

If we wish to assert we don't live in a Multiverse, then we need to
abandon the computational supervenience thesis. But that would entail
that a copy of a conscious program need not be conscious,
contradicting the so called ``Yes, doctor'' axiom of
computationalism. So we would also need to abandon computationalism.

However, if we embrace computationalism in a robust Multiversal
reality, then we are led to the conclusion that the exact form and
structure of any primitively  executing computer can have no
explanatory or causative role on empirical physics. Thus we are led to
the {\em reversal}, physics is entirely determined by the properties
of universal computation, which in turn is completely determined by
any sufficiently rich system, such as integer arithmetic.

%\bibliographystyle{plain}
%\bibliography{rus}

\appendix

\section{Measure over the universal dovetailer}

Fix a universal machine $U$, and let ${\cal P}$ be the set of all
programs of $U$.

First we need to define the equivalence class of programs that perform
an identical computation for their first $k$ steps.

\begin{definition}
$\alpha_{jk}\subset {\cal P}$ is a partitioning on ${\cal P}$, ie
$\bigcup_j\alpha_{jk}={\cal P}$ where $p\in\alpha_{jk}$ iff $\forall
x\in\alpha_{jk}$, the first $k$ steps of $p$ are counterfactually
equivalent to those of $x$.
\end{definition}

Since it is a partitioning, we are interested in the probability
measure $\mu(\alpha_{jk})$ of a program $p\in\alpha_{jk}$ eventually
being executed by a universal dovetailer, which would correspond,
under COMP, to a measure over observer moments, when suitably
restricted to conscious programs.

\begin{definition}
\begin{equation}
\mathrm{Let}\,u_p(\alpha_{jk})=\left\{
\begin{array}{ll}
1 & \mathrm{iff}\, p\, \mbox{\rm eventually
emulates a program}\, x\in\alpha_{jk}.\\
0 & \mathrm{otherwise}
\end{array}         
\right.
\end{equation}
\end{definition}

\begin{remark}
For most programs $p, u_p(\alpha_{jk})=\delta_{p\pi}, \exists
\pi\in\alpha_{jk}$, but for dovetailers and other interpreters, the
distribution differs from the Kronecker delta. In particular, for a
universal dovetailer $p$, $u_p(\alpha_{jk})=1, \forall j$.
\end{remark}

For a given universal dovetailer running on a given reference
prefix-free universal machine, the probability measure is given by summing over
programs $p\in{\cal P}$:

\begin{eqnarray}\label{firstform-measure}
\mu(\alpha_{ik}) &=& \sum_{p\in{\cal P}} 2^{-\ell(p)}
u_p(\alpha_{ik})\nonumber\\
&=& \sum_j\sum_{p\in\alpha_{jk}} 2^{-\ell(p)}u_p(\alpha_{ik})
u_p(\alpha_{ik}).
\end{eqnarray}
with the second form coming from the partitioning
$\cup_j\alpha_{jk}={\cal P}$.

For this measure to be universal for all universal dovetailers and all
reference machines, it suffices to show that this measure can be written
recursively, in terms of itself and coefficients that {\em only}
depend on the chosen partition of programs ($\{\alpha_{ik}\}$). This
demonstrates that the measure takes into account contributions from
all nested universal dovetailers, as well as any other program that
ends up executing a counterfactually equivalent program
$x\in\alpha_{ik}$, such as programs of other reference universal
machines composed of a compiler plus a program $p\in\alpha_{ik}$.

Such a decomposition is given by
\begin{equation}
\mu(\alpha_{ik}) = \sum_j \mu(\alpha_{jk}) \sum_{p\in\alpha_{jk}}
2^{-\ell(p)}u_p(\alpha_{ik}) / \sum_{p\in{\cal
P}}2^{-\ell(p)}u_p(\alpha_{jk}),
\end{equation}
which can be seen by substituting equation (\ref{firstform-measure}).

From a computationalist point of view, a measure over observer moments
will be a measure over the sets $\alpha_{ik}$ of equivalent programs
that support particular observer moments. Such a measure is not likely
to be normalisable over all observer moments, however, since each
$\alpha_{ik}$ decomposes into subsets
$\alpha_{jk+1}\subset\alpha_{ik}$, so $\sum_{m\ge
k}\sum_{j:\alpha_{jm}\subset\alpha_{ik}}\mu(\alpha_{jm})$ will be
infinite in general. In spite of not being normalisable, the relative measure
between an observer moment and it's successor
$\mu(\alpha_{jk+1})/\mu(\alpha_{ik})$ is well defined.

\end{document}